\documentclass{article}

\usepackage[preprint,nonatbib]{neurips_2025}

\usepackage[utf8]{inputenc} 
\usepackage[T1]{fontenc}    
\usepackage{hyperref}       
\usepackage{url}            
\usepackage{booktabs}       
\usepackage{amsfonts}       
\usepackage{nicefrac}       
\usepackage{microtype}      
\usepackage{xcolor}         

\usepackage{graphicx}
\usepackage[normalem]{ulem}
\useunder{\uline}{\ul}{}
\usepackage{array}
\newcolumntype{P}[1]{>{\centering\arraybackslash}p{#1}}
\usepackage{pifont}
\usepackage{multirow} 
\usepackage{caption}
\usepackage{float}

\newcommand{\cmark}{\textcolor{green!60!black}{\ding{51}}} 
\newcommand{\xmark}{\textcolor{red}{\ding{55}}}            

\title{IRLBench: A Multi-modal, Culturally Grounded, Parallel Irish-English Benchmark for Open-Ended LLM Reasoning Evaluation}

\author{%
  Khanh-Tung Tran \\
  University College Cork\\
  Cork, Ireland \\
  \texttt{123128577@umail.ucc.ie} \\
  \And
  Barry O'Sullivan \\
  University College Cork \\
  Cork, Ireland \\
  \texttt{b.osullivan@cs.ucc.ie} \\
  \And
  Hoang D. Nguyen \\
  University College Cork \\
  Cork, Ireland \\
  \texttt{hn@cs.ucc.ie} \\
}

\begin{document}

\maketitle

\begin{abstract}
    Recent advances in Large Language Models (LLMs) have demonstrated promising knowledge and reasoning abilities, yet their performance in multilingual and low-resource settings remains underexplored. Existing benchmarks often exhibit cultural bias, restrict evaluation to text-only, rely on multiple-choice formats, and, more importantly, are limited for extremely low-resource languages. To address these gaps, we introduce IRLBench, presented in parallel English and Irish, which is considered definitely endangered by UNESCO. Our benchmark consists of 12 representative subjects developed from the 2024 Irish Leaving Certificate exams, enabling fine-grained analysis of model capabilities across domains. By framing the task as long-form generation and leveraging the official marking scheme, it does not only support a comprehensive evaluation of correctness but also language fidelity. Our extensive experiments of leading closed-source and open-source LLMs reveal a persistent performance gap between English and Irish, in which models produce valid Irish responses less than 80\% of the time, and answer correctly 55.8\% of the time compared to 76.2\% in English for the best-performing model. We release IRLBench\footnote{\url{https://huggingface.co/datasets/ReliableAI/IRLBench}} and an accompanying evaluation codebase\footnote{\url{https://github.com/ReML-AI/IRLBench}} to enable future research on robust, culturally aware multilingual AI development.
\end{abstract}

\section{Introduction}

    \begin{figure}
      \centering
      \includegraphics[width=0.85\linewidth]{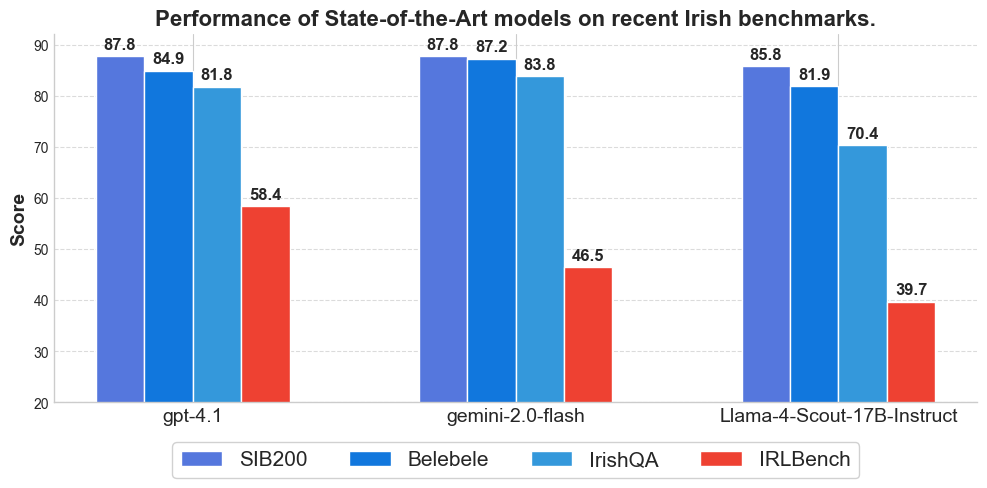}
      \caption{Performance of State-of-the-Art models on recent Irish benchmarks compared to IRLBench. We benchmark more models on IRLBench in Section~\ref{sec:exp}.}
      \label{fig:irish_data_comparison}
    \end{figure}

    Large Language Models (LLMs) and their variants, including Vision–Language Models (VLMs) and Large Reasoning Models (LRMs), have recently achieved remarkable reasoning performance \cite{cherian2024evaluating,huang-chang-2023-towards}, reshaping the field of natural language processing and rapidly overtaking traditional benchmarks \cite{wu2025bitterlessonlearned2000}. As capabilities have advanced, existing reasoning benchmarks, such as mathematical and STEM reasoning, once considered challenging, have become saturated, prompting the community to develop more rigorous and advanced evaluations to probe the limits of current models~\cite{sun2025challengingboundariesreasoningolympiadlevel,phan2025humanitysexam}. However, these benchmarking efforts predominantly focus on resource-rich and dominant languages such as English. Consequently, performance in other languages remains poorly understood, and the unique challenges of low-resource settings remain under-explored \cite{ghosh2025multilingualmindsurvey,tran2025scalingtesttimecomputelowresource}.

    Multilingual benchmarks that do exist often suffer from several shortcomings. They may embed cultural bias, limit themselves to text-only questions, rely on multiple-choice formats, or omit extremely low-resourced languages. This leads to known-limitations such as memorization problem and unable to test language generation capability in multi-choice benchmark. 
    Moreover, the translation of English-centric benchmarks into other languages often proves insufficient, as the resulting tasks may not correlate effectively with local human preferences and knowledge contexts~\cite{adilazuarda-etal-2024-towards,wu2025bitterlessonlearned2000}.
    A key example of such a gap is Irish, classified as definitely endangered by UNESCO~\cite{Unesco2010-st}. Irish currently lacks detailed, open-ended evaluation datasets. Existing Irish datasets (e.g., SIB200~\cite{adelani-etal-2024-sib}, Belebele~\cite{bandarkar-etal-2024-belebele}, IrishQA~\cite{tran2024uccix}) focus on more traditional tasks (e.g., topic classification), and are quickly saturated by state-of-the-art models, with accuracy rates approaching 90\%, leaving little room for meaningful comparison or deeper analysis of multilingual transfer, as shown in Figure \ref{fig:irish_data_comparison}.

\begin{figure}
      \centering
      \includegraphics[width=0.85\linewidth]{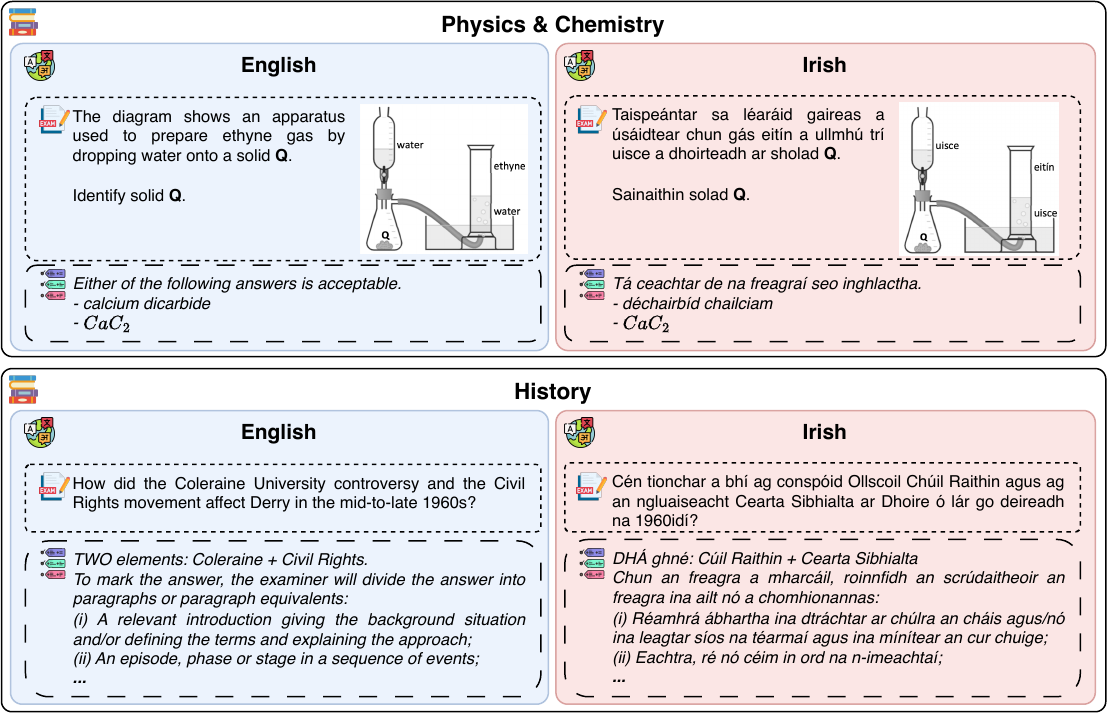}
      \caption{Illustrative examples showcasing the Irish-English parallel nature and the diverse tasks of IRLBench dataset.}
      \label{fig:irlbench_example}
    \end{figure}

    To address these issues, we introduce IRLBench, a novel multimodal, culturally grounded, parallel English–Irish benchmark for open-ended evaluation. Inspired by recent studies leveraging educational materials as benchmarks~\cite{pawar2024surveyculturalawarenesslanguage}, IRLBench is derived directly from the 2024 Irish Leaving Certificate exams. It comprises twelve representative subjects grouped into four key domains: Science, Applied Science, Business Studies, and Social Studies. To construct the dataset, we collected official exam papers and corresponding marking schemes in PDF format, employing an automated vision-language pipeline to extract and format content efficiently. A final human verification step ensures high data quality and accuracy. Because each exam is available in parallel English and Irish versions, with identical content, IRLBench naturally supports direct multilingual comparisons. By formulating each question as a long-form, open-ended generation task aligned with official marking criteria, the benchmark enables comprehensive evaluation of factual correctness, domain-specific academic knowledge, linguistic fidelity, and depth of reasoning simultaneously in both languages (Figure~\ref{fig:irlbench_example}). Furthermore, IRLBench includes multimodal questions, requiring models to interpret figures, tables, and diagrams alongside textual descriptions, as shown in Figure~\ref{fig:irlbench_example}, where a Chemistry question necessitates understanding of an associated diagram depicting a chemical process.

    We conduct extensive experiments on leading closed-source (e.g., gpt-4.1, gemini-2.0-flash) and open-source (e.g., Llama-4-Scout-Instruct) models. Using LLM-as-a-judge for automatic evaluation \cite{zheng2023judging,ye2025justice}, our results reveal a persistent performance gap between English and Irish: the best model achieves 76.2\% accuracy in English but only 55.8\% in Irish, and produces responses in Irish less than 80\% of the time. These findings underscore both the challenge of open-ended academic reasoning in low- and extremely low-resource languages, and the importance of multimodal, culturally grounded evaluation.

    Our key contributions are as follows:
    \begin{itemize}
        \item \textbf{IRLBench}. We introduce IRLBench, a parallel Irish-English, multimodal, open-ended academic reasoning dataset drawn from the 2024 Irish Leaving Certificate, 
        \item \textbf{Comprehensive Evaluation}. We systematically evaluate a range of closed- and open-source LLMs, analyzing their performance across language (English vs. Irish) and domain (Science, Applied Science, Business Studies, and Social Studies).
        \item \textbf{Insights on Multilingual Transfer}. Our experiments reveal a significant English-Irish performance gap (upto 20\% for SoTA models), quantify models’ language fidelity (where models struggling to generate output in target language), and identify subject areas where multilingual transfer is particularly weak, providing potential directions for future research on low-resource, open-ended reasoning.
    \end{itemize}
    
        

    \begin{table}[]
    \centering
    \resizebox{1.0\linewidth}{!}{
    \begin{tabular}{@{}lllllll@{}}
    \toprule
    \textbf{Dataset} & \textbf{Multilingual} & \textbf{Parallel} & \textbf{Culture aware} & \textbf{Multi-task} & \textbf{Multi-modal} & \textbf{Evaluation} \\ \midrule
    MGSM~\cite{shi2023language}           & \cmark & \cmark & \xmark  & \xmark  & \xmark  & multi-choice              \\
    Global\_MMLU~\cite{singh2025globalmmluunderstandingaddressing}   & \cmark & \cmark & \cmark & \cmark & \xmark  & multi-choice              \\
    MMMLU          & \cmark & \cmark & \xmark  & \cmark & \xmark  & multi-choice              \\
    M3Exam~\cite{zhang2023mexam}         & \cmark & \xmark  & \cmark & \cmark & \cmark & multi-choice              \\
    HLE~\cite{phan2025humanitysexam}            & \xmark  & \xmark  & \xmark  & \cmark & \cmark & open-ended \\
    \hline
    IRLBench (OURs) & \cmark & \cmark & \cmark & \cmark & \cmark & open-ended \\ \bottomrule
    \end{tabular}
    }
    \caption{Comparison of IRLBench and other related low-resource scenario benchmarks.}
    \label{tab:lit_comparison}
    \end{table}

\section{Related Works}
    \textbf{Multilingual Reasoning benchmark.} Recent multilingual reasoning datasets have attempted to evaluate LLMs across multiple languages. For instance, MGSM~\cite{shi2023language}, a multilingual variant of GSM8K~\cite{cobbe2021gsm8k}, provides human-translated arithmetic word problems in 10 languages, including two extremely low-resource languages. However, these problems predominantly reflect U.S.-centric contexts, introducing potential cultural biases. Another notable benchmark, MMLU (Massive Multitask Language Understanding)~\cite{hendryckstest2021}, has inspired multilingual derivatives: MMMLU\footnote{\url{https://huggingface.co/datasets/openai/MMMLU}}, a professionally translated set of MMLU questions across 14 languages, and Global\_MMLU~\cite{singh2025globalmmluunderstandingaddressing}, which highlights cultural biases by identifying that approximately 28\% of original MMLU questions heavily rely on Western-centric concepts. To mitigate this, Global\_MMLU categorizes its dataset into culturally sensitive and culturally agnostic subsets across 42 languages. Further, there have been efforts to create language-specific variants of MMLU using local exam questions and educational materials. Examples include CMMLU (Chinese)\cite{li-etal-2024-cmmlu}, KMMLU (Korean)\cite{son-etal-2025-kmmlu}, and ArabicMMLU (Arabic)\cite{koto-etal-2024-arabicmmlu}. Recently, M3Exam addressed cultural bias by collecting real-world exam questions from various countries, but its non-parallel structure across languages limits fair cross-lingual comparisons. Moreover, all these aforementioned benchmarks primarily use multiple-choice or short-answer questions, testing factual recall rather than in-depth generative reasoning. Such tasks risk memorization biases and fail to rigorously assess the language generation capabilities of models\cite{salido2025othersgeneraltechniquedistinguish}.

    Recent English-centric datasets, including SPIQA~\cite{pramanick2024spiqa} and Humanity’s Exam~\cite{phan2025humanitysexam}, have transitioned towards open-ended, free-form generative evaluation. However, there is still a need for such datasets for extremely low-resource languages. Our work extends this direction by introducing IRLBench, the first multilingual, culturally grounded, parallel benchmark specifically designed for open-ended generative evaluation in an extremely low-resource language scenario, Irish. Table~\ref{tab:lit_comparison} provides a comparative summary highlighting IRLBench's unique features relative to existing multilingual benchmarks.

    \textbf{Benchmarks for the Irish Language.} Irish, classified as definitely endangered by UNESCO, significantly lacks resources for AI and natural language processing development. Current datasets, such as IrishQA~\cite{tran2024uccix} or multilingual benchmarks including SIB200~\cite{adelani-etal-2024-sib} and Belebele~\cite{adelani-etal-2024-sib}, primarily target language understanding tasks such as topic classification. These datasets have been rapidly saturated by SoTA LLMs, achieving accuracies approaching 90\% (as shown in Figure~\ref{fig:irish_data_comparison}), which limits their effectiveness in differentiating model capabilities and analyzing deeper aspects of multilingual reasoning and transfer. IRLBench aims to fill this critical gap by providing challenging, multimodal, open-ended tasks that comprehensively evaluate the generative reasoning abilities of models in both Irish and English.

    \begin{figure}
      \centering
      \includegraphics[width=0.65\linewidth]{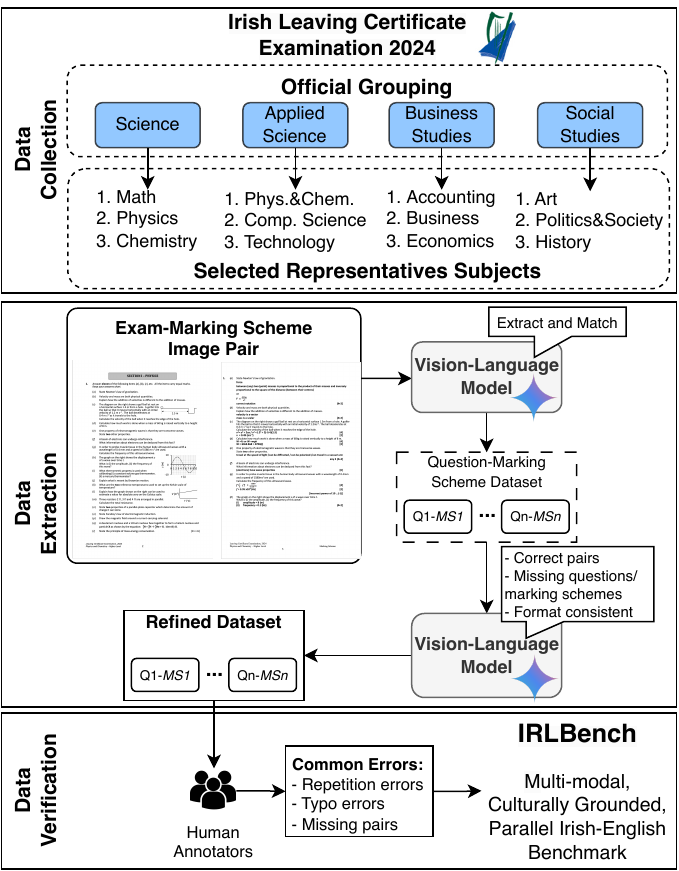}
      \caption{Pipeline.}
      \label{fig:collection_pipeline}
    \end{figure}

\section{IRLBench}

    In this section, we provide a detailed description of IRLBench, covering our data collection methodology, dataset composition, and benchmarking task designed to evaluate SoTA LLMs. First, we outline our automated extraction pipeline and subsequent human verification process. Next, we present the dataset's composition, organized by subjects and subject groups. Finally, we discuss our proposed evaluation procedure, highlighting how IRLBench effectively assesses models' open-ended generative abilities in extremely low-resource multilingual scenarios.

    \subsection{Data Collection}
        The Irish Leaving Certificate is Ireland’s final secondary-school system examination, which also serves as the standard university matriculation examination, typically requiring students at least two years of preparation\footnote{\url{https://en.wikipedia.org/wiki/Leaving_Certificate\_\%28Ireland\%29}}. Exams are available in parallel Irish and English versions, accompanied by standardized marking schemes, and cover five broad subject groups: Languages, Natural Sciences, Applied Sciences, Social Sciences, and Business Studies. Each subject can be examined at one of three difficulty levels: Higher (informally called Honours), Ordinary (Pass), or Foundation, with Higher Level exams posing the most difficult challenge. To rigorously evaluate SoTA LLMs, IRLBench exclusively includes Higher Level exams. Additionally, to facilitate direct performance comparisons between English and Irish and to avoid confounding effects from language-specific content, we exclude the Languages subject group, which is only available in one language. From each remaining group, we select three representative subjects, forming a diverse set of 12 subjects. Each subject’s official marking scheme enables precise and structured evaluation. The structure and characteristics of the selected exams are illustrated in the upper portion of Figure \ref{fig:collection_pipeline}, and example question-marking scheme pairs are provided in Figure~\ref{fig:irlbench_example}, demonstrating the multi-modal, parallel, and diverse nature of IRLBench.

        The original exam papers and marking schemes are publicly available as PDFs containing complex layouts with embedded tables, figures, and equations\footnote{\url{https://www.studyclix.ie}}. Given the challenges in accurately converting these documents into editable formats (e.g., Markdown or Word), we process them directly as images. To achieve efficient and accurate extraction, we leverage the frontier VLM gemini-2.0-flash, chosen for its strong performance in agentic tasks and cost-effectiveness~\cite{gemini2,gemini2flash}. Specifically, the model extracts pairs of exam questions and corresponding marking schemes from the PDF images, aligning textual and graphical elements, as illustrated in the lower part of Figure \ref{fig:collection_pipeline}.

        The initial extraction result reveal several systematic issues, including incorrect question-marking scheme pairings, incomplete graphic extraction, and inconsistent formatting (in table and equations). To address these challenges before human annotation, we implement an automated verification step using the same VLM. This step significantly reduce systemic errors by detecting misalignments and extraction inaccuracies. Then, human annotators is only involved in a final verification phase, manually resolving the three most common remaining issues: duplicated outputs (primarily from lengthy Irish text segments), typographical mistakes, and missing question-scheme pairs. The resulting benchmark, IRLBench, is a multimodal, culturally grounded dataset with parallel Irish-English samples, enabling detailed evaluation of multilingual academic reasoning capabilities.


    \subsection{Data Statistics}

        IRLBench consists of 1700 samples, each containing a pair of question and marking scheme developed from the 2024 Irish Leaving Certificate exams. Samples are equally distributed between English and Irish, with parallel content for comparative evaluation. Additionally, 18.11\% of these samples contain graphical illustrations, enabling multi-modal assessment.

        Table~\ref{tab:data_statistics} summarizes the dataset distribution across four subject groups: Science, Applied Science, Business Studies, and Social Studies. The Applied Science group has the highest number of samples (590), primarily from Physics and Chemistry (370). Science follows closely with 5982 samples, with Chemistry (256) and Physics (216) comprising the majority. Business Studies contains 326 samples, evenly split between Business and Economics, but with fewer samples from Accounting. Social Studies is the smallest group, but still consists of 202 samples, distributed among History (106), Politics and Society (56), and Art (40).

        Figure~\ref{fig:subject_distribution} visualizes the proportional distribution of these samples, showing the dominance of Applied Science and Science, highlighting their substantial representation within IRLBench. This structured diversity allows fine-grained analyses across subjects and language modalities, providing a robust foundation for evaluating multilingual and multimodal generative capabilities.

        \noindent
        \begin{minipage}[t]{0.4\textwidth}
        \captionsetup{type=table}
        \captionof{table}{Amount of samples collected for IRLBench. Each sample is a pair of question and marking scheme, including both text and illustrations in the original Leaving Certificate exam. 18.11\% of the samples include graphical illustrations, and the amount of samples are equally split between English and Irish, with the same task contents.}
        \centering
        \resizebox{1.37\textwidth}{!}{
        \begin{tabular}{@{}p{0.9cm}p{1.7cm}P{1cm}|p{0.9cm}p{1.6cm}P{1.2cm}@{}}
        \toprule
        Group                            & Subject               & Samples & Group                             & Subject  & Samples \\ \midrule
        \multirow{4}{*}{Science}         & Math                  & 110                  & \multirow{4}{*}{\shortstack[l]{Business\\Studies}} & Business & 132                  \\
         & Physics          & 216 &  & Accounting           & 62 \\
         & Chemistry        & 256 &  & Economics            & 132 \\ \cline{2-3} \cline{5-6} 
         & Total            & 582 &  & Total                & 326 \\ \hline
        \multirow{4}{*}{\shortstack[l]{Applied\\Science}} & Physics and Chemistry & 370                 & \multirow{4}{*}{\shortstack[l]{Social\\Studies}}   & Art      & 40                  \\
         & Computer Science & 100 &  & Politics and Society & 56 \\
         & Technology       & 120 &  & History              & 106 \\ \cline{2-3} \cline{5-6} 
         & Total            & 590 &  & Total                & 202 \\ \bottomrule
         \multicolumn{6}{c}{\textbf{Total samples across all groups: 1700}} \\
        \end{tabular}
        \label{tab:data_statistics}
        }
        \end{minipage}
        \hfill
        \begin{minipage}[t]{0.41\textwidth}
        \captionsetup{type=figure}
        \centering
        \includegraphics[width=\textwidth]{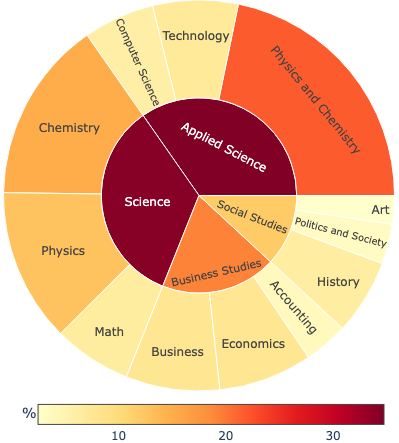}
        \captionof{figure}{Distributions of subject groups and subjects as percentages of total amount of samples in IRLBench.}
        \label{fig:subject_distribution}
        \end{minipage}

    \subsection{Benchmarking Task}
        Evaluating open-ended generation tasks poses significant challenges, as answers that are correct often vary widely in wording and detail, making traditional metrics like BLEU~\cite{papineni-etal-2002-bleu} and ROUGE~\cite{lin-2004-rouge} inadequate. Recent works~\cite{zhou2023lima,zheng2023judging,ye2025justice} address these limitations by leveraging instruction-tuned LLMs as automated evaluators, utilizing their inherent ability to assess natural language outputs in context. Following this approach of LLM-as-a-judge, we employ an evaluation strategy combined with official marking schemes to reliably assess performance on IRLBench. We select gemini-2.5-flash~\cite{gemini2} as our evaluation model due to its strong performance in instruction-following and reasoning tasks. For each evaluation, the judge model receives a prompt containing the original exam question, its official marking scheme, and the candidate model's generated response. The judge then assigns a binary label: correct or incorrect, simplifying the evaluation task and reducing the inherent variability associated with LLM-based evaluation. Although this binary judgement deviates somewhat from the granular scoring outlined in official marking schemes, it ensures consistent and comparable performance metrics across benchmarked models.

        In addition to evaluating correctness, we assess the capability of models to generate text in Irish, an extremely low-resource language. To this end, we use a FastText-based language identification model \cite{joulin2016bag,joulin2016fasttext} to determine the language of responses generated for Irish-language questions. As the language detection model operates at the sentence level, we split generated answers into individual sentences. If more than 50\% of sentences are detected as English rather than Irish, we classify the entire response as non-Irish. This criterion provides an estimate of language fidelity, emphasizing models' practical ability to handle extremely low-resource multilingual generation tasks.

        Finally, we benchmark and compare prominent SoTA VLMs. This includes leading closed-source commercial models such as o4-mini~\cite{o4-mini}, gpt-4.1~\cite{gp41}, and gemini-2.0-flash, known for their superior performance across various reasoning tasks, as well as open-source models, notably Llama-4-Scout-Instruct~\cite{llama4} and the aya-vision series (32B and 8B parameters)~\cite{dash2025ayavisionadvancingfrontier}.

\section{Results and Analysis}
    \label{sec:exp}

    \subsection{Performance Gap Between Irish and English}
\begin{figure}
          \centering
          \includegraphics[width=1.0\linewidth]{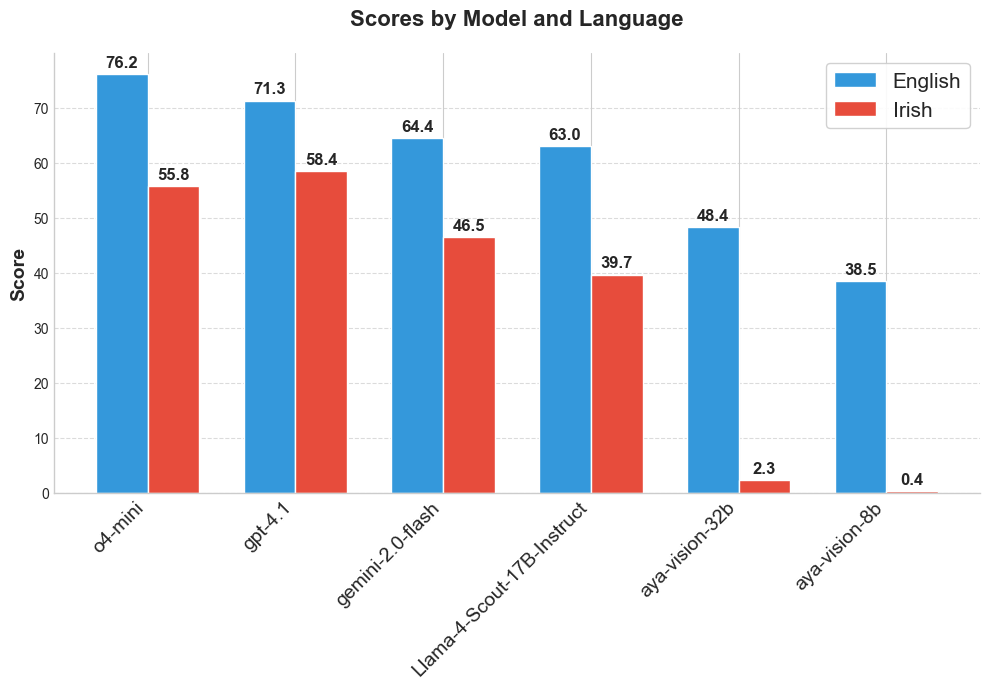}
          \caption{Accuracy scores on IRLBench per model and language.}
          \label{fig:score_lang_model}
        \end{figure}

        Figure~\ref{fig:score_lang_model} compares the accuracy scores of various models evaluated on IRLBench, segmented by language (English vs. Irish). Our results highlight that open-ended reasoning remains substantially more challenging in extremely low-resource languages such as Irish. On the English split, the top-performing model, o4-mini, achieves a strong average accuracy of 76.2\%, closely followed by gpt-4.1 at 71.3\%. In contrast, all evaluated models demonstrate considerably lower accuracy on the Irish version, scoring below 60\% despite being presented with the same questions but written in Irish. This indicates a persistent performance gap exceeding 10\% between the two languages across all models. 

        Generally, models with superior performance on English also perform relatively better on Irish. However, reasoning-adapted models (e.g., through reinforcement learning with test-time scaling), such as o4-mini, exhibit a notably larger gap (20.4\%) compared to more general models like gpt-4.1 (12.9\%) and gemini-2.0-flash (17.9\%). This suggests that current SoTA models, particularly those optimized explicitly for reasoning, have not adequately addressed multilingual transfer, resulting in pronounced performance disparities in low-resource language contexts.

    \subsection{Performance Across Subject Groups}

        Table~\ref{tab:score_per_group} summarizes model accuracy across IRLBench's subject groups, providing detailed comparisons between English and Irish performance. On the English subset, o4-mini achieves the highest overall accuracy (76.17\%), followed by gpt-4.1 (71.29\%). In general, open-source models lag behind their closed-source counterparts, particularly in science-focused categories (Science and Applied Science). However, an interesting exception occurs in the Social Studies group, where the open-source model Llama-4-Scout-Instruct outperforms the closed-source Gemini-2.0-Flash by a margin of 1.98\%.

        Noteably, performance gaps are even more significant in Irish, with smaller open-source models (aya-vision-32b and aya-vision-8b) failing almost entirely. Additionally, a consistent trend emerges: models excelling in English tend to perform better in Irish as well, underscoring that underlying reasoning proficiency strongly influences multilingual transfer. This result emphasizes the importance of robust linguistic and reasoning abilities in achieving strong multilingual generalization.

        \begin{table}[]
        \centering
        \begin{tabular}{@{}l|P{1.0cm}|P{1.4cm}P{1.4cm}P{1.4cm}P{1.4cm}|P{1.4cm}@{}}
        \toprule
        \textbf{Model} & \textbf{Split} & \textbf{Science} & \textbf{Applied Science} & \textbf{Business Studies} & \textbf{Social Studies} & \textbf{Average} \\ \midrule
        aya-vision-8b & \multirow{6}{*}{English} & 32.31 & 36.77 & 37.43 & 63.37 & 38.55 \\
        aya-vision-32b & & 42.76 & 47.11 & 44.17 & 71.29 & 48.36 \\
        Llama-4-Scout-Instruct &  & 61.26 & 57.91 & 60.73 & {\ul 78.22} & 62.99 \\
        gemini-2.0-flash & & 62.77 & 61.83 & 60.12 & {76.24} & 64.41 \\
        o4-mini & & \textbf{77.06} & \textbf{73.35} & \textbf{71.16} & \textbf{81.19} & \textbf{76.17} \\
        gpt-4.1 & & {\ul 72.21} & {\ul 68.44} & {\ul 68.10} & 72.28 & {\ul 71.29} \\ \hline
        aya-vision-8b & \multirow{6}{*}{Irish} & 0.22 & 0.00 & 0.62 & 1.00 & 0.36 \\
        aya-vision-32b &  & 1.77 & 1.00 & 0.62 & 7.92 & 2.27 \\
        Llama-4-Scout-Instruct &  & 34.10 & 39.31 & 44.98 & 49.50 & 39.67 \\
        gemini-2.0-flash & & 45.35 & 38.08 & 47.85 & 67.32 & 46.54 \\
        o4-mini & & {\ul 51.58} & {\ul 49.87} & {\ul 53.99} & {\ul 72.28} & {\ul 55.82} \\
        gpt-4.1 & & \textbf{54.12} & \textbf{54.31} & \textbf{60.12} & \textbf{73.27} & \textbf{58.43} \\ \bottomrule
        \end{tabular}
        \caption{Model performances per subject group and language of IRLBench.}
        \label{tab:score_per_group}
        \end{table}

    \subsection{Error Analysis: Correctness vs. Language Fidelity}
        Beyond correctness, we also evaluate the ability of models to generate text in Irish, an extremely low-resource language. We employ a FastText-based language identification model \cite{joulin2016fasttext} trained specifically for language detection to classify each response at the sentence level. Responses containing more than 50\% English sentences are considered non-Irish, representing a lower bound for language fidelity.

        \begin{figure}
          \centering
          \includegraphics[width=1.0\linewidth]{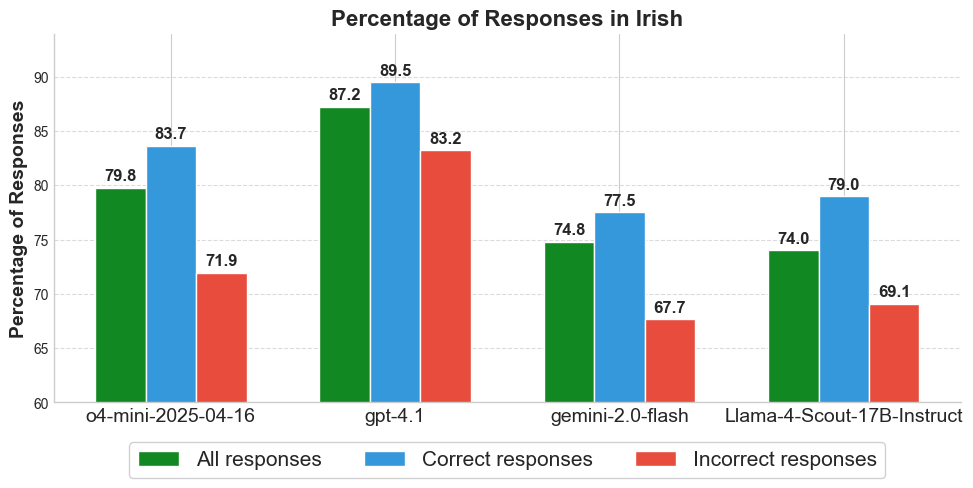}
          \caption{Percentage of responses generated by models that are in Irish (on Irish split of IRLBench).}
          \label{fig:response_in_irish}
        \end{figure}

        Figure~\ref{fig:response_in_irish} illustrates the proportion of responses produced in Irish, split by categories: for all responses, in correct responses only, or in incorrect responses only. We exclude models with extremely low accuracy (shown in previous results) due to insufficient valid samples for analysis. The analysis reveals that most evaluated models (with the exception of gpt-4.1) produce valid Irish responses less than 80\% of the time, underscoring significant gaps in generative proficiency. Moreover, there is a clear correlation between correctness and the capability to generate outputs in Irish (The percentage in Correct responses compared to the percentage in Incorrect responses), a gap of at least 6.3\% in gpt-4.1, suggesting interdependence between reasoning and language-generation skills. This result also highlights a notable limitation of previous benchmarks, which often rely on multiple-choice formats and thus do not adequately measure generative language capabilities.

        \begin{figure}
          \centering
          \includegraphics[width=0.9\linewidth]{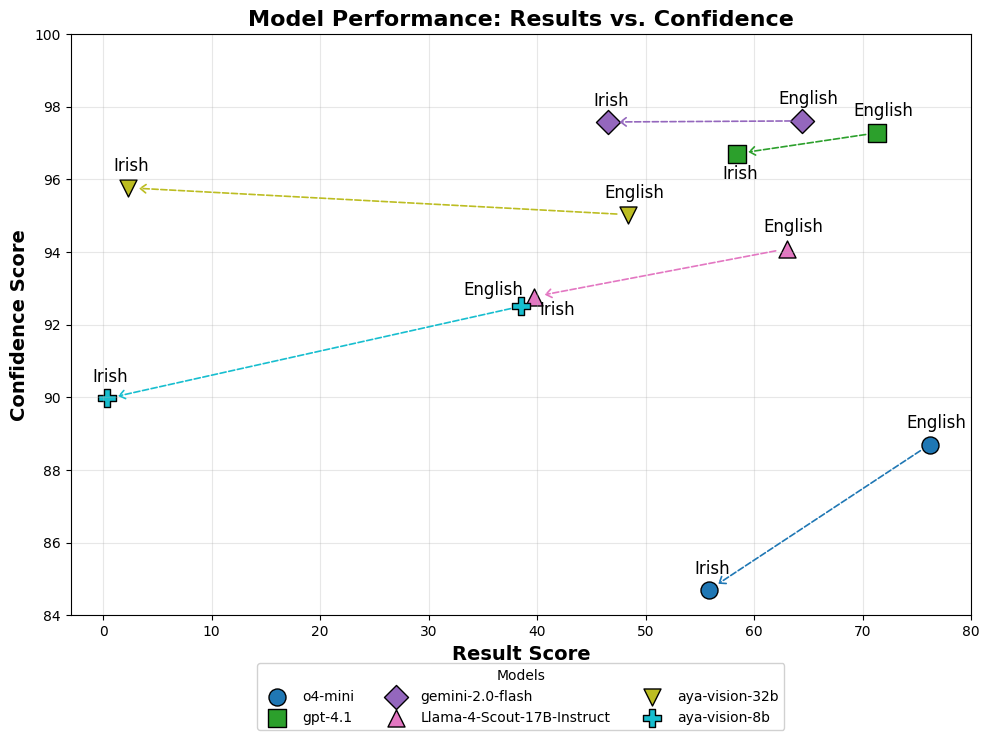}
          \caption{Model self-reported confidences compared to performances~\cite{wei2024measuringshortformfactualitylarge}.}
          \label{fig:score_confidence}
        \end{figure}
    \subsection{Confidence Analysis}
        Finally, following the methodology proposed in \cite{wei2024measuringshortformfactualitylarge}, we analyze models' self-reported confidence alongside their responses, by directly prompting them to output their confidence, alongside their response to the task. Figure~\ref{fig:score_confidence} compares actual model accuracy against their self-assessed confidence scores. Ideally, well-calibrated models should demonstrate a close alignment between confidence and accuracy (e.g., 90\% confidence corresponding to 90\% accuracy). However, all models evaluated on IRLBench display substantial discrepancies, exhibiting overly high confidence, all above 80\%, relative to actual performance where they all achieve less than 60\% score on Irish split. For example, aya-vision-32b maintains relatively high confidence, above 90\%, and even increased for Irish tassk, despite an accuracy rate of only 2.3\%,  a significant drop exceeding 40\% from English to Irish. Such miscalibration could mislead users who rely on confidence scores to assess response reliability. This poor calibration likely arises from models' intrinsic next-token prediction mechanisms, prone to hallucinations and overly optimistic self-assessments.

\section{Conclusion}
    We introduce IRLBench, the first multilingual, multi-modal benchmark specifically designed for open-ended generative evaluation in extremely low-resource scenario. As we detail our extraction pipeline from the 2024 Irish Leaving Certificate examination, IRLBench enables an in-depth assessment of multilingual transferability and academic reasoning capabilities of SoTA LLMs in both Irish and English. Our systematic evaluations reveal substantial performance gaps between the two languages (more than 10\% across all models), underscoring the significant challenges that persist in multilingual reasoning tasks, particularly in resource-limited contexts.

    \textbf{Limitations and Societal Impact.} Our benchmark currently addresses a single extremely low-resource language scenario (Irish). Furthermore, the current evaluation methodology leverages an LLM-as-a-judge paradigm, which may present challenges in scalability and robustness. Despite these limitations, the clear performance disparities identified by IRLBench highlight critical gaps in current LLMs, emphasizing the importance of developing robust multilingual generative capabilities. By releasing IRLBench, we hope to foster future research towards improving model performance and linguistic diversity in natural language processing.

    \textbf{Disclaimer.} The data used throughout this research is based on the Regulations on Re-use of Public Sector Information Regulations 2005 (SI 279 of 2005), and should be used for non-commercial research and study only.

\newpage

\bibliographystyle{ieeetr}
\bibliography{custom}

\end{document}